\def\year{2020}\relax
\documentclass[letterpaper]{article} 
\usepackage{aaai20}  
\usepackage{times}  
\usepackage{helvet} 
\usepackage{courier}  
\usepackage[hyphens]{url}  
\usepackage{graphicx} 
\usepackage{graphicx}  
\usepackage{amsmath}
\usepackage{array,multirow,graphicx}
\usepackage{framed}

\title{Multimodal Dialogue State Tracking By QA Approach 
\\with Data Augmentation}
\author{
\textbf{
    Xiangyang Mou\textsuperscript{\rm 1}, 
    Brandyn Sigouin\textsuperscript{\rm 1}, 
    Ian Steenstra\textsuperscript{\rm 1}, 
    Hui Su\textsuperscript{\rm 1,2}
}\\ 
\textsuperscript{\rm 1}Rensselaer Polytechnic Institute, 110 Eighth Street, Troy, NY 12180 USA\\ 
\textsuperscript{\rm 2}IBM Thomas J Watson Research Center, 1101 Kitchawan Road, Yorktown Heights, NY 10598 USA\\
\{moux4, sigoub, steeni\}@rpi.edu \\
huisuibmres@us.ibm.com
}

\begin{document}

\maketitle

\begin{abstract}
Recently, a more challenging state tracking task, Audio-Video Scene-Aware Dialogue (AVSD), is catching an increasing amount of attention among researchers. Different from purely text-based dialogue state tracking, the dialogue in AVSD contains a sequence of question-answer pairs about a video and the final answer to the given question requires additional understanding of the video. This paper interprets the AVSD task from an open-domain Question Answering (QA) point of view and proposes a multimodal open-domain QA system to deal with the problem. The proposed QA system uses common encoder-decoder framework with multimodal fusion and attention. Teacher forcing is applied to train a natural language generator. We also propose a new data augmentation approach specifically under QA assumption. Our experiments show that our model and techniques bring significant improvements over the baseline model on the DSTC7-AVSD dataset and demonstrate the potentials of our data augmentation techniques.
\end{abstract}

\section{Introduction}
Given a conversation flow, question-answering dialog systems are an ideal mechanism for investigating the nuances of dialog state-tracking. This is based on the hypothesis that the natural language response to any question depends on the point in time in the conversation that the question is asked. A simple example of this is if one asks, ``is there a cat in the video?''. One may ask a natural follow-up question such as ``what color is it?'', where the subject of the question is a pronoun, but the true meaning of which is stored in the dialog around the question instead of directly within it. The logical response to this may depend on such information. However, without considering the previous question(s), it is difficult for a generative model to produce information about the subject as there is little to no relevant context from which to deduce this. Even if there is information about ``color'' in the video modality, the word ``it'' is still ambiguous without understanding the current state of the dialog.  To capture such a relationship as it pertains to natural language generation, we investigate dialog history encoding techniques in order to fuse text with the other modalities. We believe that by successfully answering questions in the Audio Visual Scene-aware Dialog track of DSTC8, it will provide evidence that dialog context in QA setting does in fact store information pertinent to the current stage of a conversation.

Furthermore, we introduce a new data augmentation technique for dialog state-tracking problems. We believe that for QA problems, the presence of information in a dialog matters more than its temporal location. For each question in the dialog, we encode the QA pairs in the dialog history up until the point of the question we aim to answer. Leveraging this claim, we shuffle the dialog histories effectively increasing the size of our dataset. Additionally, we find that teacher forcing during the training procedure is important for modeling natural language sentence generation. Unlike our argument with dialog state tracking, we believe that there is a time dependency present within a sentence. During generation, the current word or token being predicted depends on the word(s) generated previously and the respective order. It should be noted that in the case of the first word, we use the special start-of-sentence token as the default first dependency. To capture this relationship during training, each token prediction must be treated as its own event; thus, the model should assume that the previously generated words are correct.

\section{Background}
Dialog State Tracking aims to model natural language by leveraging the argument that conversational patterns maintain information. Thus, there should be contextual information hidden in the dialog around the current conversation point. One of the key domains of interest for this proposal is that of Dialog Question Answering systems. In common conversations, questions asked later than the first utterance likely pertain to earlier utterances in some way, such as follow-up questions. Realistically, these relationships are not always clear and it is challenging to evaluate if there is indeed a dependency from utterance to utterance. Alternatively, one must consider the possibility that, in these types of conversations, utterances may be singleton disjoint events. This argument is important for extending encoder-decoder architectures to dialog systems. Originally used for natural language translation, these models rely on the formation of a context vector from sentence components, where a time dependency most likely exists \cite{cho2014learning}. Intuitively, if one is to extend this theory to the macro-level of full conversation, it is natural to assume that a similar relationship must exist. Question Answering systems provide a measurable way to evaluate these dialog state tracking hypotheses.

The Audio Visual Scene-Aware Dialog track of the Dialog Systems Technology Challenge 8 (AVSD, DSTC8) exists to encourage further research into the complex domain of natural language generation from multimodal data. DSTC8 and the AVSD track are an extension of DSTC7. This investigation is based on data from DSTC7. The provided modalities in the challenge dataset include visual data in the form of processed video frames, the audio data extracted from those videos, and three text modalities consisting of a summary, caption, and dialog history. The dialog history is comprised of ten question-answer pairs per example. The challenge participant is free to use any or all of the modalities, but is encouraged to attempt synthesis with the text and video-derived inputs. Ideally, the challenge aims to fuse computational linguistics, computer vision, and signal processing to generate meaningful natural language. The development of this technology is important for the emerging fields of human-agent interaction beyond just the scope of language to language interactions. One may envision an agent which aids a user through interactions with visual data. The user should be able to freely inquire about that information while expecting a reasonably confident response. This work as a whole largely extends the encoder-decoder model used for natural language translation tasks \cite{cho2014learning}. However, instead of just deriving a context vector from language alone, AVSD encourages using context from a much larger scope, which presents the unique challenge of multimodal fusion.

\section{Related Work}
Dialog State Tracking, as a research domain, is broadly defined as the deduction of evidence from linguistic information over the course of a conversation to complete a task \cite{williams-etal-2013-dialog}. Within the context of AVSD, the state of the dialog is part of the internal representation of the data from which sentence generation is based upon. Essentially the goal of state tracking is the maintenance a belief state \cite{mrkvsic2016neural}. These belief states act as a probability space from which a natural language generator, in our case a decoder, derives its context. Related topics include, but are not limited to, image captioning \cite{You_2016_CVPR} and, to an extent, the visual-question answer challenge \cite{Antol_2015_ICCV}.  Both of these topics cover the fusion of Computer Vision with Natural Language Understanding to generate new information. Key distinctions between VQA and AVSD is that VQA does not incorporate a dialog flow necessary for state-tracking and does not require the generation of language. A unique aspect of AVSD that is important to emphasize is that it is an open-domain QA setting and is not restricted to goal-specific tasks.

Other works that were extremely influential in the development of this model are the developments of the Gated-Recurrent Unit (GRU) and attention mechanisms \cite{bahdanau2014neural}. Bahdanau, Cho, and Bengio demonstrate the effectiveness of Recurrent Neural Networks with fewer parameters and how to address the issue of capturing meaningful temporal information within a sequence of natural language tokens. Sanabria, Palaskar and Metze \cite{sanabria2019cmu}, extended this technology to the previous iteration of the AVSD challenge to encode text modalities. The use of GRUs is shown to effectively address the issues pertaining to the larger size of traditional LSTM encoders, which can become very costly when used within multimodal models. Furthermore, they use the idea of attention to calculate the importance of a data frame on the overall goal of the encoding. By extending this idea to multimodal data, one can tie together diverse modalities via a common attention source. In theory, this should reduce the complexity of learning from just raw modality fusion.

\section{Proposed Approach}

\begin{figure*}[ht]
    \centering
    \includegraphics[width=\textwidth]{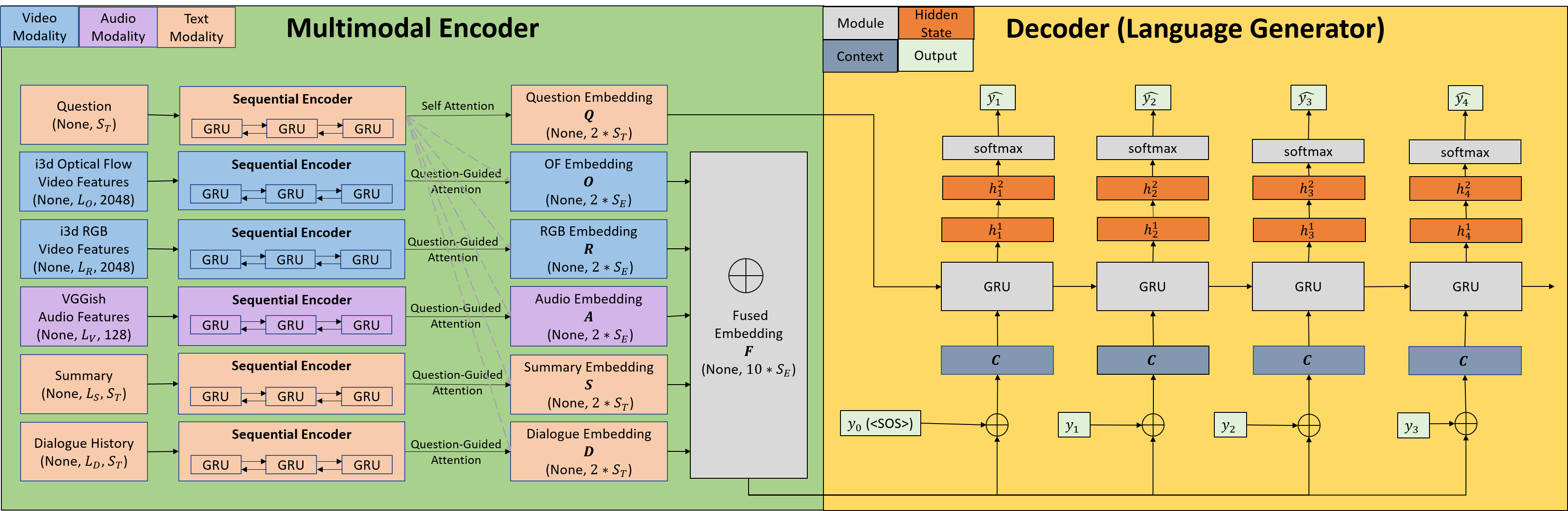}
    \caption{Model Architecture. The number and notation in the brackets, e.g. (None, $L_s$, $S_T$), describe the feature dimension. $\bigoplus$ means simple concatenation among different modalities. $y_i$ represents the $i$-th word in the ground truth.}
    \label{fig:model}
\end{figure*}

In general, our model follows an encoder-decoder framework (Fig.\ref{fig:model}) which can be commonly seen in language generation tasks \cite{wen2015semantically,wu2016google}. In encoding, bidirectional Gated Recurrent Units (BiGRU) are used for visual, audible and textual sequence embedding of which are further masked by question-guided attention. Early multimodal fusion, among different modalities, is performed to form the context representation for the decoder. The decoder takes in the context and question information in order to generate a response to the given question using a GRU. In addition, a scheduled sampling strategy \cite{bengio2015scheduled} is applied within the training phase in an effort to increase the efficiency of the training and robustness of the inference.

\subsection{Feature Encoding With Soft Attention}

Originally, in AVSD, a total of 7 different features are provided for each sample, including optical flow of video, RGB frames of video, audio, captioning, annotator generated summary, dialogue history and the question \cite{alamri2018audio}. Empirically, the best result is usually achieved by an optimal combination of features. In our work, the caption is not used because much of the information in this modality is duplicative of information found within other text modalities, such as the summary. 

For textual inputs including question, summary and individual sentences in dialogue history, we choose a pre-trained fastText model \cite{mikolov2018advances} for word embedding. We find the fact that there are a fair number of typos including missing and reversed letters within individual words caused by annotators during data collection. The typos would generate out-of-vocabulary (OOV) words and mislead the essential meaning of the sentence. The fastText embedding features a character-level encoding and is therefore considered a more suitable language model for AVSD in terms of minimizing the negative effect of OOV words in language modelling. In our work, we take the advantage of an existing library \footnote{https://github.com/plasticityai/magnitude} based on \cite{patel2018magnitude} due to its OOV handling. It would find the closest known word vector to replace an OOV word and to the best extent, restore the sentence-level representation.

For question embedding, we input the sequence of word vectors into a BiGRU, apply a 2-layer convolutional self-attention mask to the outputs of BiGRU and then take the average of the attended BiGRU outputs to obtain the final question representation.
Let $\textbf{q} = \left[q_1, q_2, \dots, q_n \right] \in R^{n \times d_w} $ denote a sequence of question word vectors with a length of $n$; where $q_i \in R^{1 \times d_w}$ is the $i$-th word vector with a dimension of $d_w$:

\begin{equation}
    \tilde{\textbf{q}} = BiGRU(\textbf{q})
\end{equation}
\begin{equation}
    m_q = ReLU(Conv2D(ReLU(Conv2D( \tilde{\textbf{q}} )))
\end{equation}
\begin{equation}
    \textbf{Q} = ReLU(
        mean( \tilde{\textbf{q}} \odot m_q)
    )
\end{equation}
where $\tilde{\textbf{q}} \in R^{n \times D}$ are the outputs of all BiGRU cells; $m_q \in R^{n \times D}$ is the attention weight; $\textbf{Q} \in R^{1 \times D}$ is the final question sentence representation; $D$ is the dimension of the output of each BiGRU cell; $Conv2D$ represents a 2-dimensional convolution layer with a size of $1 \times 1$ and $\odot$ indicates element-wise product.

For summary sentence embedding, we would like it to focus more on question-related words; therefore, we choose to use question-guided general attention \cite{luong2015effective} rather than self-attention. Then, instead of using the output of the last BiGRU cell, we do max pooling over the outputs of all BiGRU cells to form their final representations out of our belief that the embedding from max pooling includes the dominant signals across all dimensions. Let $\textbf{s} = \left[s_1, s_2, \dots, s_n \right] \in R^{n \times d_w} $ denote a sequence of summary word vectors with a length of $n$; where $s_i \in R^{1 \times d_w}$ is the $i$-th word vector with the dimension of $d_w$:
\begin{equation} \label{eq:general_attention_1}
    \tilde{\textbf{s}} = BiGRU(\textbf{s})
\end{equation}
\begin{equation}\label{eq:general_attention_2}
    m_s = softmax( \tilde{\textbf{s}} W_s \tilde{\textbf{q}}^T )
\end{equation}
\begin{equation}\label{eq:general_attention_3}
     \textbf{S} = 
     MaxPool(
        ReLU(
            \left[ m_s^T \tilde{\textbf{s}}; \tilde{\textbf{q}} \right] W_{os}
        )
     )
\end{equation}
where $\tilde{\textbf{s}} \in R^{n_s \times D}$ are the outputs of all BiGRU cells; $W_s \in R^{D \times D}$ is the trainable weight; $attn \in R^{n_s \times n_q}$ is the question-guided attention with $n_s$ as length of summary and $n_q$ as length of question; $W_o \in R^{2D \times D}$ is another trainable weight; $\textbf{S} \in R^{1 \times D}$ is the final summary sentence representation and $D$ is the dimension of the output of each BiGRU cell.

Similarly, for dialogue history, the same question-guided attention is applied following Eq.\ref{eq:general_attention_1}, Eq.\ref{eq:general_attention_2} and Eq.\ref{eq:general_attention_3}. The only difference is that, instead of using the sequence of word vectors, we use the sequence of sentence vectors to encode dialogue history. Each question and answer is treated as a single sentence and has a single sentence representation vector regardless of the actual number of sentences it contains. If no dialogue history is available for a specific question, a zero vector would be used as the representation whose elements are all zeros:
\begin{equation}
     \begin{cases}
      \textbf{d} = [q_1; a_1; \dots q_{n-1}; a_{n-1}],
      & n > 1 \\
      \textbf{d} = \textbf{0},  
      & n = 1 \\
     \end{cases}
\end{equation}
\begin{equation}
    \tilde{\textbf{d}} = BiGRU(\textbf{d}) 
\end{equation}
\begin{equation}
    m_d = softmax( \tilde{\textbf{d}} W_d \tilde{\textbf{q}}^T )
\end{equation}
\begin{equation}
     \begin{cases}
        \textbf{D} = MaxPool(
            ReLU(
                \left[ m_d^T \tilde{\textbf{d}}; \tilde{\textbf{q}} \right] W_o
            )
         )    , & n > 1 \\
        \textbf{D} = \textbf{0},  & n = 1 \\
     \end{cases}
\end{equation}
where $\textbf{d}$ is equivalent to $\textbf{s}$ in Eq.\ref{eq:general_attention_1}; $q_i, a_i$ are the question and answer sentence vector in dialogue history; $n$ means the $n$-th question-answer pair in the dialogue history starting from 1; \textbf{0} is a vector whose elements are zeros; $\tilde{\textbf{d}} = BiGRU(\textbf{d})$; $attn = softmax( \tilde{\textbf{d}} W_d \tilde{\textbf{q}}^T )$.

For video and audio modalities, we do not train our own video feature extractor but directly use the features provided by the AVSD dataset, namely i3d-flow, i3d-rgb and VGGish. i3d-flow and i3d-rgb are generated by the state-of-the-art video feature extractor  \cite{carreira2017quo} and VGGish by the state-of-the-art audio feature extractor \cite{hershey2017cnn}. Since they are all frame-wise and of variable length, we use another BiGRU to capture the temporal dependency on top of individual modality. Following the same processing procedure as shown in Eq.\ref{eq:general_attention_1}, Eq.\ref{eq:general_attention_2} and Eq.\ref{eq:general_attention_3}, we define:
\begin{equation}
    \textbf{o} = \left[ o_1; o_2; \dots o_l \right]
\end{equation}
\begin{equation}
    \textbf{r} = \left[ r_1; r_2; \dots r_m \right]
\end{equation}
\begin{equation}
    \textbf{a} = \left[ a_1; a_2; \dots a_n \right]
\end{equation}
where $\textbf{o}$, $\textbf{r}$, $\textbf{a}$ are equivalent to $\textbf{s}$ in Eq.\ref{eq:general_attention_1} and their corresponding outputs are denoted as $\tilde{\textbf{o}}$, $\tilde{\textbf{r}}$, $\tilde{\textbf{a}}$; $o_i$, $r_i$, $a_i$ denote individual frame representation for i3d-flow, i3d-rgb and audio respectively. The final i3d-flow, i3d-rgb and audio representation, are denoted as $\textbf{O}, \textbf{R}, \textbf{A}$, s.t:
\begin{equation}
     \textbf{O} = 
     MaxPool(
        ReLU(
            \left[ m_o^T \tilde{\textbf{o}}; \tilde{\textbf{q}} \right] W_{oo}
        )
     )
\end{equation}
\begin{equation}
     \textbf{R} = 
     MaxPool(
        ReLU(
            \left[ m_r^T \tilde{\textbf{r}}; \tilde{\textbf{q}} \right] W_{or}
        )
     )
\end{equation}
\begin{equation}
     \textbf{A} = 
     MaxPool(
        ReLU(
            \left[ m_a^T \tilde{\textbf{a}}; \tilde{\textbf{q}} \right] W_{oa}
        )
     )
\end{equation}
where $W_{oo}$, $W_{or}$, $W_{oa}$ are the trainable weights; $m_o^T$, $m_r^T$, $m_a^T$ are the question-guided attention masks for i3d-flow, i3d-rgb and audio modalities, following the same strategy as in Eq.\ref{eq:general_attention_2}

\subsection{Multimodal Fusion}

The context vector contains information from different modalities and will be used for natural answer generation. We form it by simple concatenation in order to achieve early fusion across multimodalities, i.e.

\begin{equation}
    \textbf{C} = 
    \left[
        \textbf{O}; 
        \textbf{R}; 
        \textbf{A}; 
        \textbf{S}; 
        \textbf{D}
    \right]
\end{equation}

\subsection{Decoder}
Because our system is open-domain and supposed to generate answers of free-form, any extraction-based language generation approach \cite{wang2016machine} would be out of our consideration. In our work, we adopt a two-layer BiGRU as the natural language generator. One good point of a RNN is that it can take in variable-length input and also generate variable-length output. More importantly, RNN is known for its ability of modelling long-term spatial or temporal dependency within a sequence so that the language generated could be more fluent and readable. A two-RNN-layer at the output is a commonly-used setting in related areas such as a language generation system and image captioning. In addition, it is said to be beneficial in decreasing the opportunity of repeated words within generated language; which, is one of the difficulties in the natural language generation task. 

Given the input question, context and the preceding words, the language generator models the probability of each next word. We would like the GRU network to be able to reason over the context and predict the next work based on the question and the preceding part of the answer.
\begin{equation}
    P(\omega_1, \dots, \omega_n)
    =
    \Pi_{i=1}^n
    P(\omega_i | \omega_{0 \sim i-1} ; \textbf{C}; \textbf{Q}; )
\end{equation}
In our work, we initialize the GRU hidden state with the question vector $Q$ and take the concatenation of the context $\textbf{C}$ and the 1-gram preceding word as the input to GRU cell. We believe that the context contains more information than the question and do not want the context to forget any information along the progressive prediction procedure.
\begin{equation} \label{eq:gru_update_1}
    z_t = \sigma (
        W_z [\textbf{C}; \textbf{w}_{t-1}] + U_{z} h_{t-1}
    )
\end{equation}
\begin{equation} \label{eq:gru_update_2}
    r_t = \sigma (
        W_r [\textbf{C}; \textbf{w}_{t-1}] + U_{r} h_{t-1}
    )
\end{equation}
\begin{equation} \label{eq:gru_update_3}
    h_t = \sigma (
        W [\textbf{C}; \textbf{w}_{t-1}] + r_t \odot U h_{t-1}
    )
\end{equation}
Eq.\ref{eq:gru_update_1}, Eq.\ref{eq:gru_update_2} and Eq.\ref{eq:gru_update_3} show how our GRU cell updates its hidden state at time $t$. $\textbf{w}_{t-1}$ is the word vector at time $t-1$ in the prediction; $[\textbf{C}; \textbf{w}_{t-1}]$ represents the input to the GRU cell; $h_t$ and $h_{t-1}$ are the hidden states at different time; $W_z$, $W_r$, $W$, $U_z$, $U_r$, $U$ are the trainable weights in GRU cell.

\section{Experiments}

\subsection{Training}
In our training phase, we use an Adam optimizer to minimize the cross entropy error between the predicted word and the ground truth. The F1 score on the validation set is used to terminate the training procedure. To increase the training efficiency and accuracy, we use teacher forcing \cite{williams1989learning}; it uses the ground truth word to predict the next word during training. More specifically in Eq.\ref{eq:gru_update_1}, Eq.\ref{eq:gru_update_2} and Eq.\ref{eq:gru_update_3}, $\textbf{w}_{t-1} = y_{t-1}$ rather than $\hat{y}_{t-1}$ just as shown in Fig.\ref{fig:model}.  

As a comparison, we also try a scheduled sampling technique \cite{bengio2015scheduled} which introduces probability into teacher forcing. Different from a traditional teacher forcing technique that always uses the ground truth word, there is a certain probability in scheduled sampling to use the predicted word as the input to the GRU cell. \cite{bengio2015scheduled} claims that scheduled sampling could improve the generalization and robustness. However, we do not see significant improvements in our tests. Therefore, we remain to use teacher forcing for our experiments.

\subsection{Data Augmentation}
Data augmentation is a widely used technique in deep learning. Most of the time, it is of great help and can outperform the baseline significantly for data driven approaches. After examining the AVSD dataset, we find a way to enlarge the size of the training set by several orders of magnitude. The training set of AVSD, provides 10 question-answer pairs for each video.
\begin{itemize}
    \item The most \textbf{basic} way of using the training data is to treat the first 9 pairs as dialogue history and take the last question as what needs to be answered.
    \item A \textbf{quick improvement} would be treating the first $n$ pairs as dialogue history and take the $n+1$-th question as what needs to be answered. This could augment the data by 10 times.
    \item In our work, since AVSD is regarded as a question answering problem, we do not necessarily care about the sequence order of the dialogue history. Each question-answer pair is being seen as a knowledge point. With this slight difference, we can \textbf{shuffle} the first $n$ pairs for the $(n+1)^{th}$ question; ideally it should not be a problem for a human to answer the $(n+1)^{th}$  question. In other words, the dialogue history $[q_1; a_1; q_2; a_2; q_3; a_3; q_4; a_4]$ can be seen as no obvious difference from $[q_1; a_1; q_4; a_4; q_3; a_3; q_2; a_2]$, or any other order to the $(n+1)^{th}$  question. Theoretically, for a training sample whose dialogue history length is 9, we can generate $P^9_9 - 1 = 362879$ similar samples out of it, following the \textbf{shuffle} idea. Thus, the approach could enlarge the training set by tens of thousands times on average.
\end{itemize}

In our work, we take the \textbf{quick improvement} version as the baseline; assuming that most of the people would adopt such technique, and take \textbf{shuffle} version as our new data augmentation approach. In our experiments, we only double the training set considering the training time.

\begin{table*}[!ht]
  \centering
  \small
  \begin{tabular}{|c|c|c|c|c|c|c|c|c|}
    \hline
     & \multicolumn{1}{c|}{Model} & \multicolumn{1}{c|}{BLEU-1} & \multicolumn{1}{c|}{BLEU-2} & \multicolumn{1}{c|}{BLEU-3} & \multicolumn{1}{c|}{BLEU-4} & \multicolumn{1}{c|}{METEOR} & \multicolumn{1}{c|}{ROUGE-L} & \multicolumn{1}{c|}{CIDEr} \\ 
     \hline
     \multirow{8}{*}{{\rotatebox[origin=c]{90}{Video + Text}}}
     & \cite{nguyen2018film} & \textbf{0.695} & \textbf{0.553} & \textbf{0.444} & \textbf{0.36} & \textbf{0.249} & \textbf{0.544} & 0.997 \\
     & \cite{le2019end} & 0.631 & 0.491 & 0.390 & 0.315 & 0.239 & 0.509 & 0.848 \\
     & \textbf{Our Model} & 0.586 & 0.436 & 0.333 & 0.262 & 0.206 &	0.46 & 0.704 \\
     & \cite{pasunuru2019dstc7} & N/A & N/A & N/A & 0.118 & 0.150 & 0.378 & \textbf{1.158} \\
     & \cite{lin2019entropy} & 0.333 & 0.196 & 0.131 & 0.093 & 0.129 & 0.334 & 0.88 \\
     & \cite{zhuang2019investigation} & 0.29 & 0.184 & 0.125 & 0.089 & 0.121 & 0.298 & 0.8 \\
     & \cite{yeh2019reactive} & 0.237 & 0.161 & 0.116 & 0.088 & 0.121 & 0.31 & 1.015 \\
     & Baseline \cite{Hori_2019} & 0.256 & 0.161 & 0.109 & 0.078 & 0.113 & 0.277 & 0.727 \\
     \hline
     \multirow{4}{*}{{\rotatebox[origin=c]{90}{Text Only}}}
     & \cite{nguyen2018film} & \textbf{0.686} & \textbf{0.52} & \textbf{0.416} & \textbf{0.340} & 0.228 & \textbf{0.518} & 0.851 \\
     & \cite{le2019end} & 0.633 & 0.49 & 0.386 & 0.31 & \textbf{0.242} & 0.515 & \textbf{0.856} \\
     & \textbf{Our Model} & 0.631 & 0.478 & 0.37 & 0.291 & 0.224 & 0.496 & 0.789 \\
     & Baseline \cite{Hori_2019} & 0.245 & 0.152 & 0.103 & 0.073 & 0.109 & 0.271 & 0.705 \\
     \hline
  \end{tabular}
  \caption{Model Performance. Models are ranked by an overall performance rather than any single metrics.}
  \label{table:result}
\end{table*} 

\begin{table*}[!ht]
  \centering
  \small
  \begin{tabular}{|c|c|c|c|c|c|c|c|c|}
    \hline
     & \multicolumn{1}{c|}{Model} & \multicolumn{1}{c|}{BLEU-1} & \multicolumn{1}{c|}{BLEU-2} & \multicolumn{1}{c|}{BLEU-3} & \multicolumn{1}{c|}{BLEU-4} & \multicolumn{1}{c|}{METEOR} & \multicolumn{1}{c|}{ROUGE-L} & \multicolumn{1}{c|}{CIDEr} \\ 
     \hline
     \multirow{6}{*}{{\rotatebox[origin=c]{90}{Video + Text}}}
     & TF + MaxPool + BiGRU (Baseline) & 0.587 & \textbf{0.438} & 0.334 & 0.261 & 0.204 & 0.451 & 0.684 \\
     & + Data Augmentation ($\times 2$) & 0.586 & 0.436 & 0.333 & \textbf{0.262} & \textbf{0.206} & \textbf{0.46} & \textbf{0.704} \\
     & BiGRU $\rightarrow$ BiLSTM & 0.585 & 0.437 & \textbf{0.335} & 0.262 & 0.202 & 0.45 & 0.684 \\
     &  TF $\rightarrow$ SS & \textbf{0.588} & 0.437 & 0.333 & 0.259 & 0.197 & 0.449 & 0.655 \\
     & MaxPool $\rightarrow$ AveragePool& 0.378 & 0.25 & 0.165 & 0.111 & 0.152 & 0.345 & 0.3414 \\
     & - TF & 0.448 & 0.231 & 0.124 & 0.063 & 0.12	& 0.353	& 0.224 \\
     \hline
  \end{tabular}
  \caption{Ablation Study. ``TF" means ``Teacher Forcing". ``SS" means ``Scheduled Sampling". ``No-TF" means No Teacher Forcing or Scheduled Sampling. ``$\times 2$'' means ``Data Augmentation by a factor of 2".}
  \label{table:ablation}
\end{table*}

\subsection{Results}
\subsubsection{Evaluation Metrics}
We use 7 metrics for evaluating our model which are widely used when evaluating image and video captioning, as well as language generation: BLEU(1-4) \cite{Papineni:2002:BMA:1073083.1073135}, METEOR \cite{denkowski-lavie-2014-meteor}, ROUGE-L \cite{lin2004rouge}, and CIDEr \cite{DBLP:journals/corr/VedantamZP14a}.

\subsubsection{Model Performance}
In order to further evaluate our model's performance, we compare our results with the baseline model; as well as other participants in the DSTC7 challenge on the track of AVSD under two different settings -- with and without video related modalities (i3d-flow, i3d-rgb and audio). Our best model fully utilizes the combination of teacher forcing, max pooling, BiGRU and data augmentation techniques. Table.\ref{table:result} shows that our best model outperforms the baseline model significantly and our scores are comparatively better than the majority of other models, demonstrating that our model successfully captures salient signals among multimodalities. But comparing to \cite{nguyen2018film} under the Text-Only setting, whose architecture is specifically designed to encode the conversation flow, we see a big decrease in all scores. We hypothesize that as we regard the dialogue history as a group of discrete QA pairs, we miss certain inner temporal dependencies, impacting tasks like coreference resolution. While our data demonstrates that the ordering of the dialog history as a whole may not contribute as much information as previously assumed, it is important to consider that phenomena like coreference resolution most likely does depend on temporal location. Fig. \ref{fig:example_of_coreference_resolution} shows an example of the importance of deducing meanings of ambiguous terms within a dialog history. Another example is in the case of an ambiguous pronoun. Both situations usually rely on the presence of a previous noun or a certain portion in an image or video. After such information has been deduced, our data suggests that the order of that processed information would likely have little effect on the performance of the system. However, without doing this before shuffling, information that can be obtained through techniques such as coreferencing may be untapped.

\begin{figure}
    \begin{framed}
        Q1: Is there a guy or a girl in the video? \\
        A1: A man with beard.   \\
        Q2: Is that man alone?  \\
        A2: \underline{\qquad\qquad\qquad\qquad}
    \end{framed}
    \centering
    \caption{An example of dialog history with a coreference resolution issue from the AVSD dev set. If $Q_2$ was the given question, in order to offer a correct answer, a human would need to figure out whom ``that man" exactly referred to.}
    \label{fig:example_of_coreference_resolution}
\end{figure}

\begin{figure*} [!ht]
    \centering
    \includegraphics[width=\textwidth]{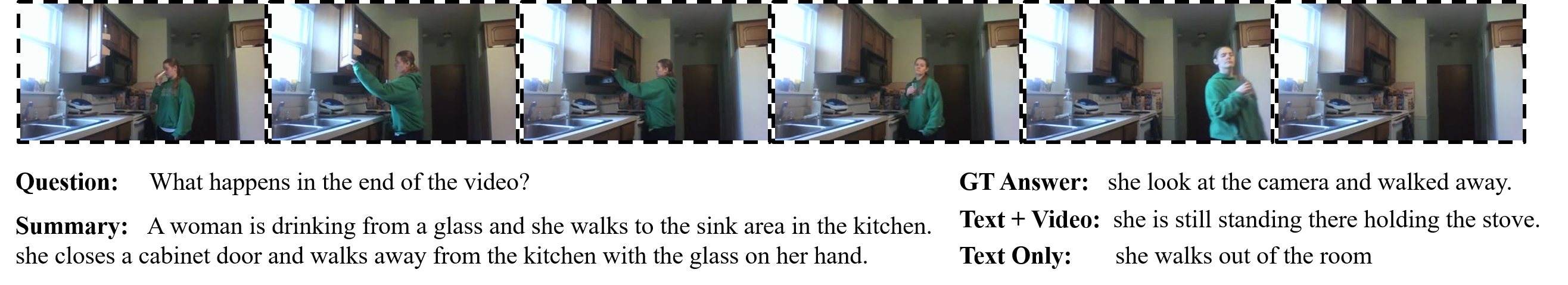}
    \caption{A question whose answer requires the understanding of dynamics in the video. But the \textbf{Text + Video} model provides an answer describing all static actions, which shows that the captured feature more focuses on the image individually and doesn't well represent the dependency among the frames. It could partially explain the performance decrease compared to the \textbf{Text Only} model.}
    \label{fig:result_samples1}
\end{figure*}

\begin{figure*} [!ht]
    \centering
    \includegraphics[width=\textwidth]{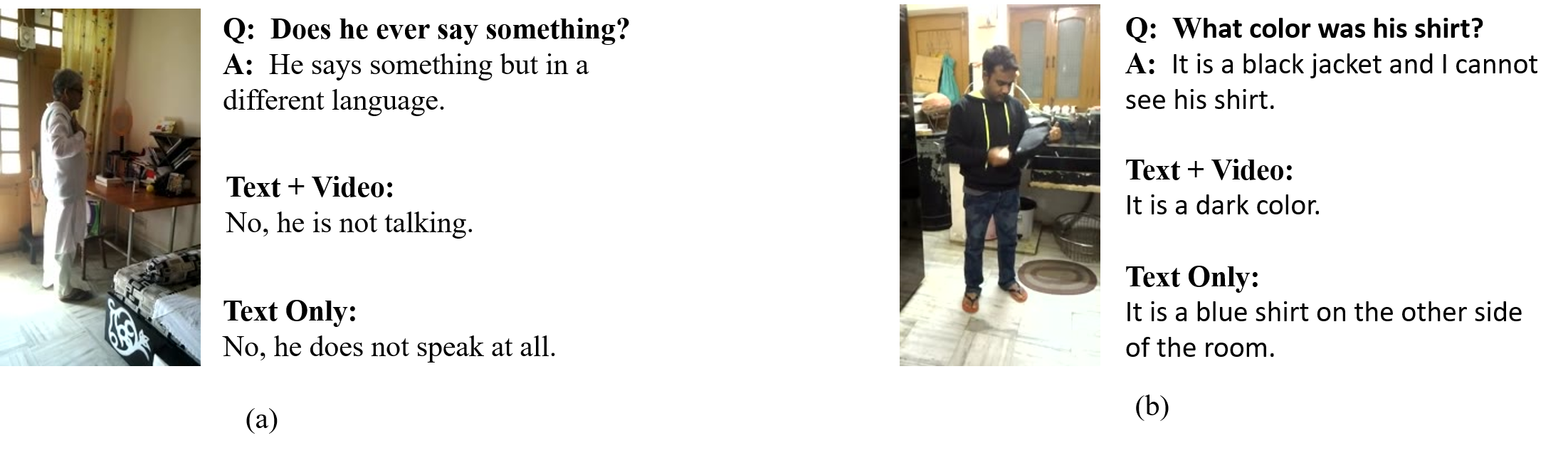}
    \caption{\textbf{A} is the ground-truth answer to the question \textbf{Q}. (a) is an instance where information from audible inputs is directly correlated with the proper response. (b) shows the necessity for proper visual-textual reasoning.}
    \label{fig:result_samples2}
\end{figure*}

When comparing our model under different settings, we notice a significant empirical improvement in every metric when shifting from Video-Text to just Text-Only. This is counter-intuitive because the scores should decrease as the context should be less-complete without the video and audible inputs. On the contrary, the two other models shown evaluating on text-only do see a slight decrease in their scores when compared to the same models that make use of the video-derived modalities. The difference could be the result of our model not capturing the real attention of the video-related signals within relevance to the current questions at the hidden layer of the BiGRU \cite{nguyen2018film}. Without the proper video encoding, the video embeddings could be simply adding noise to our model. Fig. \ref{fig:result_samples1} provides an example of such an error. This provides an interesting insight: the video encoder seems to struggle at capturing temporal actions, limiting the amount of useful information present in the encoding. Since this encoding cannot add meaningful information to the context in respect to the current question, it functionally becomes noisy data. Given such findings, the development of more robust video-audio encoding techniques would be a logical next step in this research. The success of the text-only model in this context can likely be explained by the model being able to capture relationships between keywords in the input modalities and the words that appear in the ground truths. In \ref{fig:result_samples2} example (a), the model is still able to produce a relevant answer, albeit an empirically incorrect one, regarding audio in the video without actually being able to access that modality. It is safe to conclude here that our model successfully captures textual context, however will struggle in producing the most robust responses when critical information is exclusive to other modalities. 

Fig. \ref{fig:result_samples2} provides examples for the importance of utilizing a combination of an image encoder, audio encoder and text understanding. Leaving any of these out could result in inadequate answers compared to the ground truths. For instance, example (a) shows the importance of an audio encoder. The question can only be answered correctly with the information from the audio, but the vocal attention is overshadowed by other modalities. Example (b) illustrates the importance of using image encoding with text understanding because they work together to find that the cloth is black (or dark). 

\subsubsection{Ablation Study}
We conduct our ablation study under the Video-Text setting. As shown in Table.\ref{table:ablation}, the techniques of teacher forcing, maximum pooling, average pooling, scheduled sampling, data augmentation and RNN variations has been testified. We find that the best overall scores are from the model that uses teacher forcing, maximum pooling, data augmentation and BiGRUs.

We are interested in the performance difference between BiLSTM and BiGRU since both are widely used RNN variants in others' work. \cite{weiss2018practical} claims that LSTM with ReLU activation function is strictly stronger for NLP tasks than GRU because of its unbound computational ability; however, our results share more or less identical in terms of the end-goal performance. Given that our initial model has very low CIDEr score of 0.224, we experiment by including teacher forcing. This leads to a notable increase in all our metrics by a range of around $30\% \sim 300\%$. Time wise, we find in our experiments that the model with teacher forcing needs fewer epochs to reach the same performance as without it. Because we find teacher forcing to be an improvement, we also try scheduled sampling. However, we find that it does not improve our scores beyond the improvements from strict teacher forcing. Since scheduled sampling has a certain probability to use the predicted token instead of a ground-truth token as the last token, scheduled sampling could work if our baseline model (TF + MaxPool + GRU) has a fairly high generative accuracy to begin with as there would be less uncertainty during next-token prediction. We find that using the prediction is too noisy and only makes the training procedure less efficient; as well as not being beneficial to cover a limited number of outliers in inference. We could try lowering our probability of picking the predicted word (0.2), but lowering by too much could defeat the purpose of using scheduled sampling. 

Once we switch from average pooling to maximum pooling, every evaluation metric increases dramatically, most notably the BLEU and CIDEr scores. This verifies our belief that max pooling includes the dominant signals across all dimensions from the outputs of the BiGRU cells. Lastly, with the inclusion of data augmentation with a factor of 2, our scores increase even further. Therefore, an enlarged dataset through shuffling of the $n$ pairs for the $(n+1)^{th}$ question does result in a quick score enhancement over the baseline. Additionally, this supports our theory that the information within a dialog history matters more than the order it appears in. Given this performance, we will experiment with other factors, such as 4 and 5, in order to find the extent of how much improvement can be made.

\section{Conclusion}

In this paper, we evaluate various techniques such as max pooling, use of BiGRU/BiLSTM encoders, teacher forcing/scheduled sampling, and our proposed data augmentation technique on question answering from multimodal data. Our goal was to analyze dialog state tracking through the perspective of QA within the AVSD track of the DSTC8. Through this research, we empirically conclude that our approach performs satisfactorily and improves upon the work of DSTC7. We find that increasing the dataset by a factor of 2 by shuffling dialog history, combined with teacher forcing, max pooling, and GRU encoders, produces the best results within the scope of our tests. Teacher forcing and our proposed technique of shuffling dialog histories result in a substantial improvement over the baseline model and our own tests which do not use these methods. This appears to reinforce our hypothesis that for QA problems, the order of dialog histories is less important than the raw information present within. Additionally, the success derived from the incorporation of teacher forcing suggests that individual tokens within a sentence do have a very important temporal dependence which is critical for generating accurate natural language. 

In the future, we would like to conduct further investigation into the fusion of visual and textual data. Specifically, we would like to experiment with approaches pertaining to video-derived modalities, in hope to produce more informative responses with specific details extracted from the videos.

\section*{Acknowledgment}
This work is supported by Cognitive and Immersive Systems Lab (CISL), a collaboration between IBM and RPI, and also a center in IBM’s AI Horizons Network.

\bibliographystyle{aaai}
\bibliography{reference}
\end{document}